\documentclass[letterpaper, 10 pt, journal, twoside]{IEEEtran}

\IEEEoverridecommandlockouts                              

\usepackage[utf8x]{inputenc}
\usepackage{amsmath}
\usepackage{amssymb}
\usepackage{wrapfig}
\usepackage{graphicx}
\usepackage{algorithm}
\usepackage{algpseudocode}
\usepackage{eqparbox}
\usepackage{mdwmath}
\usepackage{mdwtab}
\usepackage{array}
\usepackage{authblk}
\usepackage{bm}

\usepackage{hyperref}
\usepackage{units}
\usepackage{multirow}

\usepackage{subfigure}

\usepackage{setspace}

\usepackage[left=2cm,right=2cm,top=2cm,bottom=2cm]{geometry}

\usepackage[english]{babel}

\usepackage{pslatex}

\graphicspath{
	{figs/}
	{matlab/}
	{anim/}
}

\renewcommand{\vec}[1]{\mathbf{#1}}

\title{\LARGE \bf
NeuroTrajectory: A Neuroevolutionary Approach to Local State Trajectory Learning for Autonomous Vehicles
}

\author{Sorin~M.~Grigorescu,
        Bogdan~Trasnea,
        Liviu~Marina,
		Andrei~Vasilcoi
		and~Tiberiu~Cocias
\thanks{Manuscript received: February, 11, 2019; Revised April, 26, 2019; Accepted June, 11, 2019.}
\thanks{This paper was recommended for publication by Editor Nancy Amato upon evaluation of the Associate Editor and Reviewers' comments.}
\thanks{The authors are with Elektrobit Automotive
		and the Robotics, Vision and Control Lab (ROVIS) (\url{www.rovislab.com}),
        Transilvania University of Brasov, Romania.
        {\tt\small sorin.grigorescu@elektrobit.com}}%
\thanks{Digital Object Identifier (DOI): see top of this page.}
}

\begin{document}

\markboth{IEEE Robotics and Automation Letters. Preprint Version. Accepted June, 2019}
{Grigorescu \MakeLowercase{\textit{et al.}}: NeuroTrajectory: A Neuroevolutionary Approach to Local State Trajectory Learning} 

%

\maketitle

\begin{abstract}

Autonomous vehicles are controlled today either based on sequences of decoupled perception-planning-action operations, either based on End2End or Deep Reinforcement Learning (DRL) systems. Current deep learning solutions for autonomous driving are subject to several limitations (e.g. they estimate driving actions through a direct mapping of sensors to actuators, or require complex reward shaping methods). Although the cost function used for training can aggregate multiple weighted objectives, the gradient descent step is computed by the backpropagation algorithm using a single-objective loss. To address these issues, we introduce \textit{NeuroTrajectory}, which is a multi-objective neuroevolutionary approach to local state trajectory learning for autonomous driving, where the desired state trajectory of the ego-vehicle is estimated over a finite prediction horizon by a \textit{perception-planning} deep neural network. In comparison to DRL methods, which predict optimal actions for the upcoming sampling time, we estimate a sequence of optimal states that can be used for motion control. We propose an approach which uses genetic algorithms for training a population of deep neural networks, where each network individual is evaluated based on a multi-objective fitness vector, with the purpose of establishing a so-called \textit{Pareto front} of optimal deep neural networks. The performance of an individual is given by a fitness vector composed of three elements. Each element describes the vehicle's travel path, lateral velocity and longitudinal speed, respectively. The same network structure can be trained on synthetic, as well as on real-world data sequences. We have benchmarked our system against a baseline Dynamic Window Approach (DWA), as well as against an End2End supervised learning method.

\end{abstract}




\section{Introduction}
\label{sec:introduction}

\begin{figure}
	\centering
	\begin{center}
		\includegraphics[scale=0.96]{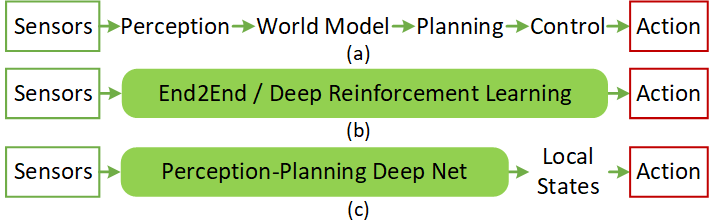}
		\vspace{-2em}
		\caption{\textbf{From a modular pipeline to a \textit{perception-planning} deep neural network approach for autonomous vehicles}. Green symbolizes learning components. (a) Mapping sensors to actuators using a traditional pipeline. The output of each module provides input to the adjoining component. (b) Monolithic deep network for direct mapping of sensory data to control actions. (c) \textit{Perception-Planning} deep neural network (our approach).}
        \label{fig:had_deep_net}
	\end{center}
	\vspace{-2.5em}
\end{figure}

An autonomous vehicle is an intelligent agent which observes its environment, makes decisions and performs actions based on these decisions. The driving functions map sensory input to control output and are often implemented as modular perception-planning-action pipelines, such as the one illustrated in Fig.~\ref{fig:had_deep_net}(a). In a modular pipeline, the main problem is divided into smaller subproblems, where each module is designed to solve a specific task. The ability of an autonomous vehicle to find a route between two points, a start location and a desired location, represents \textit{path planning}. To design a collision-free route, the vehicle should consider all possible obstacles present in the surrounding environment.

Over the last couple of years, deep learning has become a leading technology in many domains.
Research in self-driving technologies has been dominated by End2End and Deep Reinforcement Learning (DRL) methods (see Fig.~\ref{fig:had_deep_net}(b)), implemented as model-free discrete or continuous control systems. The functional safety of these systems is usually difficult to monitor, mainly due to their direct mapping of sensory inputs to actuators. Additionally, DRL algorithms tend to require complex reward shaping~\cite{Jaritz_ICRA2018}.

In this paper, we propose \textit{NeuroTrajectory}, which is a neuroevolutionary solution for the perception-planning deep neural network from Fig.~\ref{fig:had_deep_net}(c). In comparison to DRL or End2End, which are used to estimate optimal driving actions, in our work we focus on estimating an optimal local state trajectory over a finite prediction horizon. To implement motion control, the predicted states can be used as input to a model predictive controller. The design and implementation of the motion controller is out of the scope of this paper. We reformulate autonomous driving as a local state trajectory estimation problem for an artificial agent, where a Deep Neural Network (DNN) is used to predict a local ego-vehicle trajectory. The trajectory is defined as a sequence of future desired states, estimated over a finite prediction horizon. The weights of the DNN are calculated based on  a multi-objective fitness vector composed of three losses: the ego-vehicle's traveled path, lateral velocity and longitudinal speed.

NeuroTrajectory is trained based on a neuroevolutionary training method, which is a non-differentiable approach for calculating a so-called Pareto front of neural networks individuals. The Pareto front is defined within an objective space, having its dimension equal to the size of the fitness vector. The axes of the objective space are the three losses in the fitness vector. The training data is represented by paired sequences of \textit{Occupancy Grids} (OGs) and vehicle trajectory labels. An element in a trajectory sequence encodes the vehicle's position, steering angle and velocity.

The main contributions of this paper are summarized as follows: 1) we introduce a solution for estimating local driving trajectories by reformulating autonomous driving as a local state trajectory learning problem; 2) we define a deep neural network topology for predicting desired future trajectories; 3) we introduce a neuroevolutionary approach for training trajectory estimation systems based on Pareto optimization and genetic algorithms, which can be easily extended to different non-differentiable problems.

Throughout the paper, we use the following notation. The value of a variable is defined either for a single discrete time step $t$, written as superscript $<t>$, or as a discrete sequence defined in the $<t, t+k>$ time interval, where $k$ denotes the length of the sequence. For example, the value of a trajectory variable $\vec{y}$ is defined either at discrete time $t$ as $\vec{y}^{<t>}$, or within a sequence interval $\vec{Y}^{<t, t+k>}$. Vectors and matrices are indicated by bold symbols.

\section{Related Work}

Perception-Planning-Action pipelines, shown in Fig.~\ref{fig:had_deep_net}(a), are currently the main approach for controlling autonomous vehicles. The driving problem is divided into smaller subproblems, where separate components are responsible for environment perception, path planning and motion control.
Although the components are relatively easy to interpret due to their modularity, they are often constructed on manually chosen rules which are unsuitable for learning complex driving strategies.

\textit{End2End learning}~\cite{NVIDIA_End2End_Learning} systems directly map raw input data to control signals and are trained in a supervised manner. The training data, often in the form of images from a front-facing camera, is collected together with time-synchronized steering angles recorded from a human driver. A DNN is then trained to output steering commands. End2End systems are faced with the challenge of learning a very complex mapping in a single step.

\textit{Deep Reinforcement Learning} (DRL)
is a type of machine learning algorithm where agents are taught actions by interacting with their environment. The system does not have access to training data, but maximizes a cumulative reward quantifying its behavior.
The reward is used as a pseudo label by a Q-learning algorithm for training a DNN which is then used to estimate a discrete Q-value function approximating the next best driving action (e.g. turn left, turn right, accelerate, decelerate), given the current state.

The Q-learning approach has been extended to continuous action spaces based on policy gradient estimation~\cite{Lillicrap2016ContinuousCW}. The method in~\cite{Lillicrap2016ContinuousCW} describes a model-free actor-critic algorithm able to learn different continuous control tasks directly from raw pixel inputs. A model-based solution for continuous Q-learning is proposed in~\cite{GuLilSutLev16}. Although continuous control with DRL is possible, the most common strategy for DRL in autonomous driving is based on discrete control~\cite{Jaritz_ICRA2018}. The main challenge here is the training, since the agent has to explore its environment, usually through learning from collisions. Such systems, trained solely on simulated data, tend to learn a biased version of the driving environment. A solution here is to use the Inverse Reinforcement Learning paradigm, where a deep neural network can still be trained on labeled data, but in a RL fashion~\cite{Wulfmeier2016}.

A deep learning path planning algorithm for autonomous driving is proposed in~\cite{Yu2018}. The method addresses the problem of error modeling and path tracking dependencies. Another approach for path planning on an autonomous car is presented in \cite{Paxton2017}. In this case, DRL methodologies represent the basis for learning low-level control policies and task-level option polices. Finding the best path to a target location inside a square grid is analyzed in \cite{Aleksandr2018} with well-established path planning methods correlated to novel neural learning approaches.

Two multi-vehicle trajectory estimation methods using recurrent neural networks and sequence-to-sequence prediction have been proposed in~\cite{Kim_2017} and~\cite{Park_2018}, respectively. Both approaches take into consideration precise positions of the vehicles in a small occupancy grid space of $36 \times 21$ cells, where a vehicle is represented by a single cell. Deep imitative models for trajectory estimation have been proposed in~\cite{Rhinehart_2018}. However, only an obstacles-free environment is used for training and testing within the \textit{Carla} simulator.

Traditionally, DNNs are trained using differentiable methods based on single-objective cost functions. Although multiple losses can be aggregated into a single cost function via weighting, the gradient descent step in the backpropagation algorithm will adjust the network's weights based only on the single-objective loss. Weighting multiple losses also introduces additional hyperparameters (typically manually defined) required to weight each individual loss~\cite{Deb11}. Another approach to multi-objective optimization in deep learning is Multi-Task Learning (MTL)~\cite{Ruder17a}, where the objectives are given as tasks. The network model either shares its layers between weighted tasks (hard parameter sharing), or each single task is used to train a separate model (soft parameter sharing). In the latter case, the parameters between the models corresponding to the given tasks are regularized in order to encourage the parameters to be similar. As stated in~\cite{Ruder17a}, the hard parameter sharing paradigm is still pervasive for neural-network based MTL. In contrast to MTL, we use a Pareto multi-objective optimization technique which independently optimizes a population of DNNs, where the training of each individual is not influenced by the other individuals. In this way, we ensure a better exploration of the parameters space during training, while avoiding additional weighting hyperparameters.

\section{Method}
\label{sec:method}

\subsection{Problem Definition}

\begin{figure}
	\centering
	\begin{center}
		\includegraphics[scale=0.8]{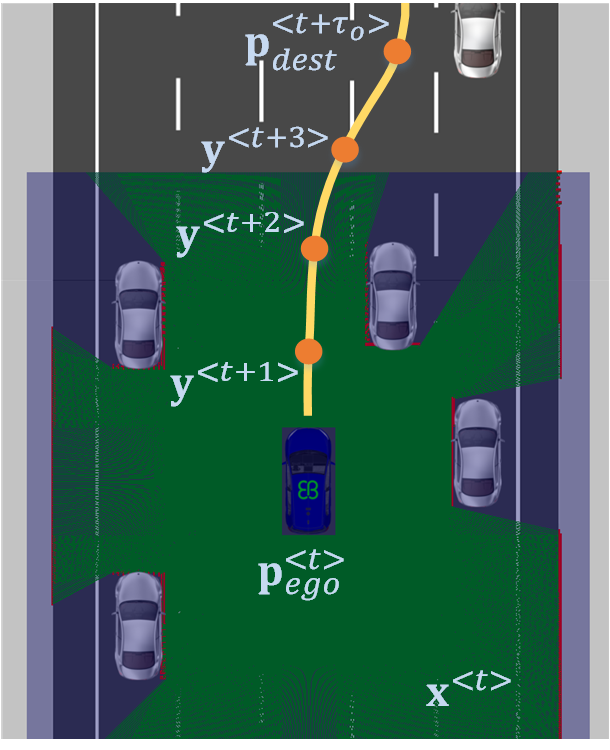}
		\vspace{-0.5em}
		\caption{\textbf{Local state trajectory estimation for autonomous driving.} Given the current position of the ego-vehicle $\vec{p}_{ego}^{<t>}$, a desired destination $\vec{p}_{dest}^{<t+\tau_o>}$ and an input sequence of occupancy grids $\vec{X}^{<t-\tau_i, t>}=[\vec{x}^{<t-\tau_i>}, ..., \vec{x}^{<t>}]$, the goal is to estimate a driving trajectory $\vec{Y}^{<t+1, t+\tau_o>}=[\vec{y}^{<t+1>}, ..., \vec{y}^{<t+\tau_o>}]$, where each element in the output sequence $\vec{Y}$ represents the desired position of the ego-vehicle at that specific moment in time.}
        \label{fig:problem_description}
	\end{center}
	\vspace{-2.3em}
\end{figure}

A basic illustration of the local state trajectory estimation problem for autonomous driving is shown in Fig.~\ref{fig:problem_description}. Given a sequence of 2D occupancy grids (OG) $\vec{X}: \mathbb{R}^2 \times \tau_i \rightarrow \mathbb{R}^2 \times \tau_o$, the position of the ego-vehicle $\vec{p}_{ego}^{<t>} \in \mathbb{R}^2$ in $\vec{x}^{<t>}$ and the destination coordinates $\vec{p}_{dest}^{<t>} \in \mathbb{R}^2$ in occupancy grid space at time $t$, the task is to learn a local trajectory for navigating the ego-vehicle to destination coordinates $\vec{p}_{dest}^{<t+\tau_o>}$. $\tau_i$ is the length of the OGs input sequence, while $\tau_o$ is the number of time steps for which the trajectory of the ego-vehicle is estimated. In other words, with $\vec{p}_{0}^{<t>}$ being a coordinate in the current OG observation $\vec{x}^{<t>}$, we seek a desired local navigation trajectory of the ego-vehicle from any arbitrary starting point $\vec{p}_0^{<t>}$ to $\vec{p}_{dest}^{<t+\tau_o>}$, with the following properties:

\begin{itemize}
	\item the traveled path $|| \vec{p}_{0}^{<t>} - \vec{p}_{dest}^{<t+\tau_o>} ||$ is minimal;
	\item the lateral velocity, given by the steering angle's rate of change $v_{\delta} \in \left[ \dot{\delta}_{min}, \dot{\delta}_{max} \right]$ is minimal, signifying a minimal value for $v_{\delta}^{<t, t+\tau_o>}$;
	\item the forward speed, also known as longitudinal velocity, $v^{<t, t+\tau_o>}$ is maximal and bounded to an acceptable range $[v_{min}, v_{max}]$.
\end{itemize}

The vehicle is modeled based on the single-track kinematic model of a robot~\cite{PadenCYYF16}, with position state $\vec{y}^{<t>} = (p^{<t>}_x, p^{<t>}_y)$ and no-slip assumptions. $p_x$ and $p_y$ represent the position of the vehicle in the 2D driving plane, respectively. The heading is not taken into consideration for trajectory estimation.

\begin{figure}
	\centering
	\begin{center}
		\includegraphics[scale=0.95]{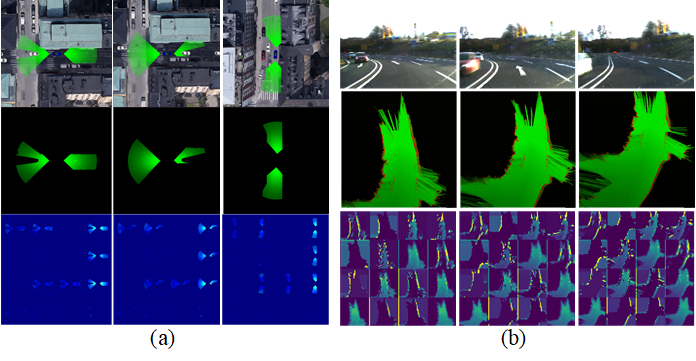}
		\vspace{-0.8em}
		\caption{\textbf{Examples of synthetic (a) GridSim and (b) real-world occupancy grids.} The top images in each group shows a snapshot of the driving environment together with its respective OG and activations of the first convolutional layer of the deep neural network in Fig.~\ref{fig:neural_network_diagram}. }
        \label{fig:occupancy_grids}
	\end{center}
	\vspace{-2.1em}
\end{figure}

We observe the driving environment using OGs constructed from fused LiDAR and radar data. Synthetic and real-world OG data samples are shown in Fig.~\ref{fig:occupancy_grids}. Green and red pixels represent free-space and obstacles, respectively, while black signifies unknown occupancy.
A single OG corresponds to an observation instance $\vec{x}^{<t>}$, while a sequence of OGs is denoted as $\vec{X}^{<t-\tau_i, t>}$. These observations are axis-aligned discrete grid sequences, acquired over time interval $[t-\tau_i, t]$ and centered on the sequence of vehicle states $\vec{Y}^{<t-\tau, t>}$. For a detailed description of OGs, please refer to~\cite{Marina_IRC2019}.

\subsection{Trajectory Estimation as a Cognitive Learning Task}
\label{sec:ba_cognitive_learning_problem}

\begin{figure*}
	\centering
	\begin{center}
		\includegraphics[scale=0.9]{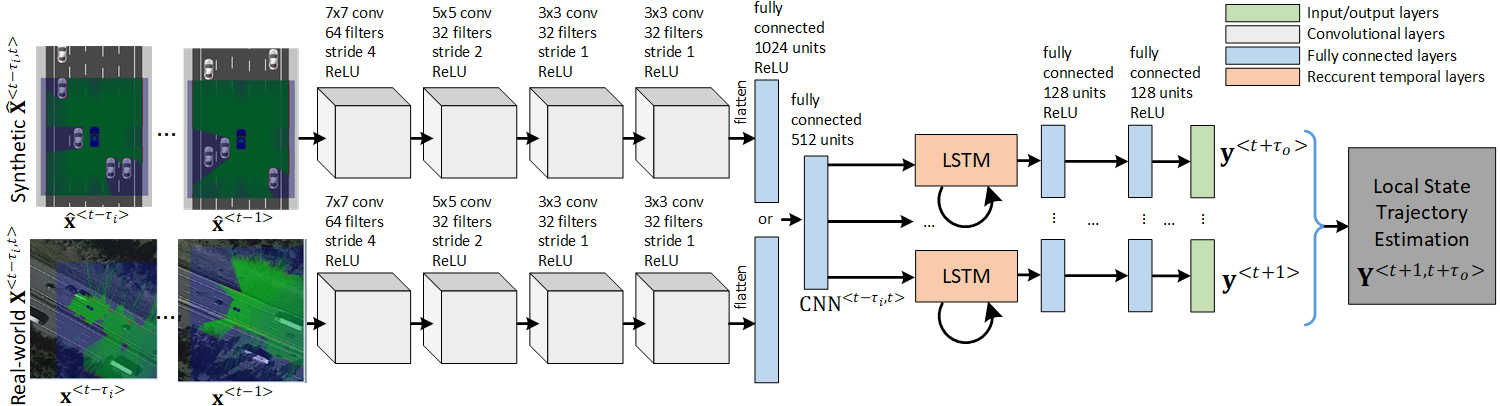}
		\vspace{-1em}
		\caption{\textbf{Deep neural network architecture for estimating local driving trajectories.} The training data and labels consists of synthetic $(\hat{\vec{X}}^{<t-\tau_i, t>}, \hat{\vec{Y}}^{<t+1, t+\tau_o>})$ or real-world $(\vec{X}^{<t-\tau_i, t>}, \vec{Y}^{<t+1, t+\tau_o>})$ OG sequences, together with their future trajectory labels. Both synthetic and real-world OG streams are passed through a convolutional neural network, followed by two fully connected layers of $1024$ and $512$ units, respectively. The obtained $CNN^{<t-\tau_i, t>}$ sequence of spatial features is further fed to a stack of LSTM branches.}
        \label{fig:neural_network_diagram}
	\end{center}
	\vspace{-2.2em}
\end{figure*}

Our approach to local state trajectory learning is to reformulate the autonomous driving problem as a cognitive learning task. The above problem can be modeled as a \textit{Markov Decision Process} (MDP) $M = (S, A, T, L)$, where:

\begin{itemize}
	\item $S$ represents a finite set of states, $\vec{s}^{<t>} \in S$ being the state of the agent at time $t$. To encode the location of the agent in the driving OG space at time $t$, we define $\vec{s}^{<t>} = \vec{X}(\vec{p}_{ego}^{<t-\tau_i, t>})$, which denotes an axis-aligned discrete grid sequence in interval $[t-\tau_i, t]$, centered on  the ego-vehicle positions' $\vec{p}_{ego}^{<t-\tau_i, t>}$.
	
	\item $A$ represents a finite set of trajectory sequences, allowing the agent to navigate through the environment, where $\vec{Y}^{<t+1, t+\tau_o>} \in A$ is the predicted trajectory that the agent should follow in the future time interval $[t+1, t+\tau_o]$. A trajectory $\vec{Y}^{<t+1, t+\tau_o>}$ is defined as a collection of estimated trajectory state set-points:
	\begin{equation}
		\begin{split}
			\vec{Y}^{<t+1, t+\tau_o>} = [\vec{y}^{<t+1>}, \vec{y}^{<t+2>}, ..., \vec{y}^{<t+\tau_o>}].
		\end{split}
		\label{eq:trajectory_set_points}
	\end{equation}
	
	\item $T: S \times A \times S \rightarrow [0, 1]$ is a stochastic transition function, where $T_{s^{<t>}, \vec{Y}^{<t+1, t+\tau_o>}}^{\vec{s}^{<t+\tau_o>}}$ describes the probability of arriving in state $\vec{s}^{<t+\tau_o>}$, after performing a motion along trajectory $\vec{Y}^{<t+1, t+\tau_o>}$.
	
	\item $\vec{L}: S \times A \times S \rightarrow \mathbb{R}^3$ is a multi-objective fitness vector function which quantifies the trajectory quality of the ego-vehicle: 
	\begin{equation}
		\vec{L}_{\vec{s}^{<t>}, \vec{Y}^{<t+1, t+\tau_o>}}^{\vec{s}^{<t+\tau_o>}} = \begin{bmatrix} l_1^{<t+\tau_o>} & l_2^{<t+\tau_o>} & l_3^{<t+\tau_o>} \end{bmatrix}.
		\label{eq:multiobjective_loss}
	\end{equation}
	
	Each element in Eq.~\ref{eq:multiobjective_loss} is defined as:
	\begin{equation}
		l_1^{<t+\tau_o>} = \sum_{i=1}^{\tau_o} || \vec{p}_{ego}^{<t+i>} - \vec{p}_{dest}^{<t+i>} ||_2^2
		\label{eq:r_1}
	\end{equation}
	\begin{equation}
		l_2^{<t+\tau_o>} = \sum_{i=1}^{\tau_o} v_{\delta}^{<t+i>}
		\label{eq:r_2}
	\end{equation}
	\begin{equation}
		l_3^{<t+\tau_o>} = \sum_{i=1}^{\tau_o} v_{f}^{<t+i>} \in [v_{min}, v_{max}]
		\label{eq:r_3}
	\end{equation}	
\end{itemize}

Intuitively, $l_1^{<t+\tau_o>}$ represents a distance-based feedback, which is smaller if the car follows a minimal energy trajectory to $\vec{p}_{dest}^{<t+\tau_o>}$ and large otherwise. $l_2^{<t+\tau_o>}$ quantifies hazardous motions and passenger discomfort by summing up the lateral velocity of the vehicle. The feedback function $l_3^{<t+\tau_o>}$ is the moving longitudinal velocity of the ego-vehicle, bounded to speeds appropriate for different road sectors, such as $v^{<t, t+\tau_o>} \in [80 kmh, 130 kmh]$ for the case of highway driving.

Considering the proposed state estimation scheme, the goal is to train an optimal approximator, defined here by a deep network, which can predict the optimal state trajectory $\vec{Y}^{<t+1, t+\tau_o>}$ of the ego-vehicle, given a sequence of occupancy grid observations $\vec{X}^{<t-\tau_i, t>}$ and the multi-objective fitness vector from Eq.~\ref{eq:multiobjective_loss}.

We learn an optimal state trajectory by combining Convolutional Neural Networks (CNN) with the robust temporal predictions of Long Short-Term Memory (LSTM) networks. The two types of neural architectures are combined as follows (see Fig.~\ref{fig:neural_network_diagram}). An observation $\vec{x}^{<t>}$ is firstly processed by a CNN, implemented as a series of convolutional layers, aiming to extract relevant spatial features from the input data. The CNN outputs a feature-space representation for each observation in $\vec{X}^{<t-\tau_i, t>}$. Each processed spatial observation in the input interval $[t-\tau_i, t]$ is flatten and passed through two fully connected layers of $1024$ and $512$ units, respectively. The input sequence into an LSTM block is represented by a sequence of spatially processed observations, denoted in Fig.~\ref{fig:neural_network_diagram} as $CNN^{<t-\tau_i, t>}$. The same network topology from Fig.~\ref{fig:neural_network_diagram} can be trained separately on synthetic, as well as on real-world data. As trainable network parameters, we consider both the weights of the LSTM networks, as well as the weights of the convolutional layers.

For computing the state trajectory of the ego-vehicle, we have designed the deep neural network from Fig.~\ref{fig:neural_network_diagram}, where OG sequences are processed by a set of convolutional layers, before being feed to different LSTM network branches. Each LSTM branch is responsible for estimating trajectory set-points along time interval $[t+1, t+\tau_o]$. The choice for a stack of LSTM branches over a single LSTM network that would predict all future state set-points comes from our experiments with different network architectures. Namely, we have observed that the performance of a single LSTM network decreases exponentially with the prediction horizon $\tau_o$. The maximum value for which we could obtain a stable trajectory using a single LSTM network was $\tau_o = 2$. As shown in the experimental results section, this is not the case with our proposed stack of LSTMs, where each branch is responsible for estimating a single state set-point. The precision difference could occur due to input-output structures of the two architectures. A single LSTM acts as a many-to-many, or sequence-to-sequence, mapping function, where the input sequence is used to generate a time-dependent output sequence. In the case of LSTM branches, the original sequence-to-sequence problem is divided into a stack of many-to-one subproblems, each LSTM branch providing a many-to-one solution, thus simplifying the search space. In our case, the solution given by a specific branch represents an optimal state set-point for a single timestamp along $[t+1, t+\tau_o]$. Another reason for the effectiveness of LSTM branches over a single LSTM might come from the actual problem addressed in this paper. That is, single LSTMs behave well in natural language processing applications, where the input and output domains are represented by text sequences, whereas the input-output to our proposed NeuroTrajectory method is represented by sequences of occupancy grids and state set-points, respectively.



\subsection{Neuroevolutionary Training for Deep Neural Networks}
\label{sec:training}

The aim of the proposed neuroevolutionary training strategy is to compute optimal weights for a collection of deep neural networks $\Phi$ by simultaneously optimizing the fitness vector $\vec{L}$ from Eq.~\ref{eq:multiobjective_loss}. $\varphi(\Theta)$ represents a single deep neural network, also called an \textit{individual}, with weights $\Theta$.

$\Phi$ represents a \textit{population} of $K$ deep networks, each network having a corresponding set of weights $\Theta$:

\begin{equation}
	\Phi =
		\begin{bmatrix}
			\varphi_1(\Theta_1), ..., \varphi_K(\Theta_K)
		\end{bmatrix}^T
\end{equation}

The fitness vector $\vec{L}$ gives a quantitative measure of the network's response, forming a multi-objective loss which can be used to tune the weights of the individuals in population $\Phi$. The training procedure does not search for a fixed set of weights, or even a finite set, but for a Pareto optimal collection of weights $\Theta^*$, where each element in the collection represents the weights of a \textit{Pareto optimal deep neural network} $\varphi^*(\Theta^*)$. The Pareto front of deep networks $\Phi^*$ is a subset of $\Phi$, defined as:

\begin{equation}
	\Phi^* =
		\begin{bmatrix}
			\varphi^*_1(\Theta^*_1), ..., \varphi^*_k(\Theta^*_k)
		\end{bmatrix}^T
\end{equation}

\noindent where $\Phi^* \in \Phi$. We coin as \textit{elite individual}, an individual located on the Pareto front.

\begin{figure}
	\centering
	\begin{center}
		\includegraphics[scale=0.95]{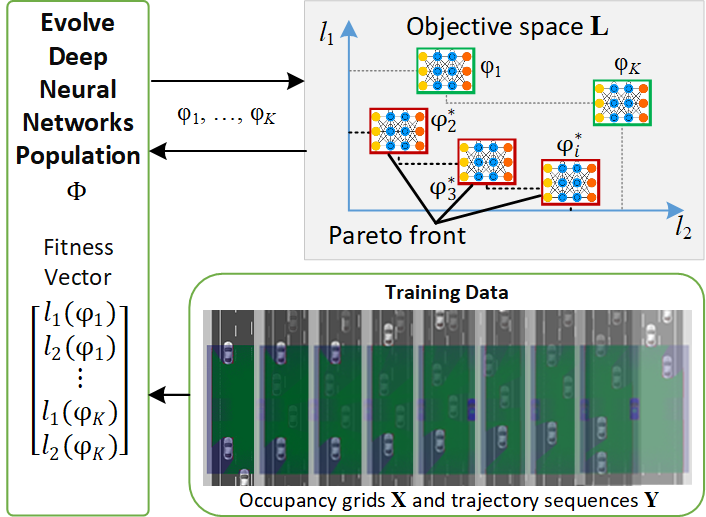}
		\vspace{-1em}
		\caption{\textbf{Neuroevolutionary training for learning local ego-vehicle state trajectories.} The training data is used to evolve a population of deep neural networks $\Phi = [\varphi_1 (\Theta_1), ..., \varphi_K(\Theta_K)]$ by learning their weights $[\Theta_1, ..., \Theta_K]$ using genetic algorithms. The training aims to optimize the multi-objective fitness vector $\vec{L} = [l_1, ..., l_W]$, in the multidimensional objective space $\vec{L}$, where each coordinate axis represents a fitness value. In this illustration, $\vec{L}$ is composed of two fitness values, $l_1$ and $l_2$. The best performing networks $\varphi^* (\Theta^*)$ lie on the so-called Pareto front in objective space.}
        \label{fig:block_diagram}
	\end{center}
	\vspace{-2.3em}
\end{figure}

The weights of a single individual are stored in a so-called \textit{solution vector} $\Theta = \begin{bmatrix} \theta_1, \theta_2, ..., \theta_n, \end{bmatrix}^T$, composed of $n$ decision variables $\theta_i$, with $i = 1, ..., n$ and $\theta \in \mathbb{R}^n$. $\theta_i$ represents a weight parameter in a single neural network. The neuroevolution training principle is illustrated in Fig.~\ref{fig:block_diagram}.

The multi-objective optimization problem takes into account a number of functional values which have to be either minimized or maximized, subject to a number of possible constraints that any feasible solution must satisfy~\cite{Deb11}:

\begin{equation}
	\begin{aligned}
		\text{minimize / maximize } & l_w(\Theta), && w = 1, 2, ..., W \\
		\text{subject to } & r_v (\Theta) \geq 0, && v = 1, 2, ..., V \\
		& h_q (\Theta) = 0, && q = 1, 2, ..., Q \\
		& \theta_i^{(L)} \leq \theta_i \leq \theta_i^{(U)}, && i = 1, 2, ..., n
	\end{aligned}
	\label{eq:multiobjective_optimization}
\end{equation}

\noindent where $l_w(\Theta)$ is the $w$-th objective function, with $w = 1, ..., W$, and $W \geq 2$ is the number of objectives. $r_v (\Theta)$ and $h_q (\Theta)$ are constraints set on the optimization process. $\theta_i^{(L)}$ and $\theta_i^{(U)}$ are lower and upper variable constraint bounds set on each decision variable, respectively.

\begin{figure}
	\centering
	\begin{center}
		\includegraphics[scale=0.9]{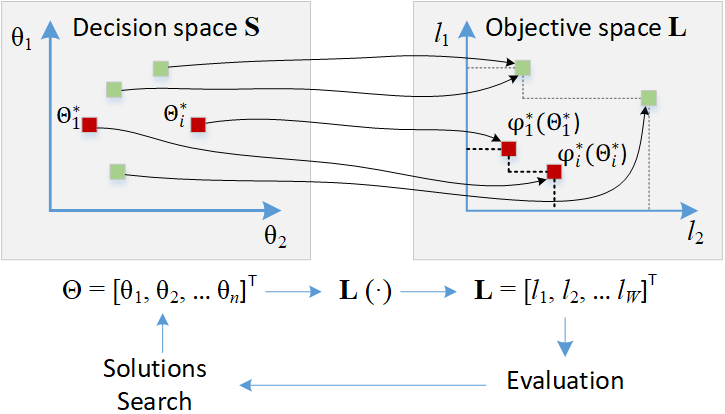}
		\vspace{-0.8em}
		\caption{\textbf{Mapping of solution vectors $\Theta$ from the \textit{decision space} $\vec{S}$ to \textit{objective space} $\vec{L}$.} Each solution $\Theta$ in decision space corresponds to a coordinate in objective space. The red marked coordinates are the set of Pareto optimal solutions $\Theta^*$ for a multi-objective minimization problem, located on the Pareto front drawn with thick black line.}
        \label{fig:pareto_optimization_problem}
	\end{center}
	\vspace{-2.3em}
\end{figure}

The solutions satisfying $r_v (\Theta)$, $h_q (\Theta)$ and the variable bounds form the so-called \textit{feasible decision variable space} $\vec{S} \in \mathbb{R}^n$, or simply \textit{decision space}. A core difference between single and multi-objective optimization is that, in the latter case, the objective functions make up a $W$-dimensional space entitled \textit{objective space} $\vec{L} \in \mathbb{R}^W$. A visual illustration of a 2D decision and objective space is shown in Fig.~\ref{fig:pareto_optimization_problem}. For each solution $\Theta$ in decision variable space, there exists a coordinate in objective space.

In Pareto optimization, there exists a set of optimal solutions $\Theta^*$, none of them usually minimizing, or maximizing, all objective functions simultaneously. Optimal solutions are called \textit{Pareto optimal}, meaning that they cannot be improved in any of the objectives without degrading at least one objective. A feasible solution $\Theta_1$ is said to Pareto dominate another solution $\Theta_2$ if:

\vspace{3mm}
\begin{enumerate}
	\item $l_i (\Theta_1) \leq l_i (\Theta_2)$ for all $i \in \{1, 2, ..., W\}$ and
	\vspace{2mm}
	\item $l_j (\Theta_1) < l_j (\Theta_2)$ for at least one index $j \in \{1, 2, ..., W\}$.
\end{enumerate}
\vspace{3mm}

A solution $\Theta^*$ is called Pareto optimal if there is no other solution that dominates it. The set of Pareto optimal solutions is entitled \textit{Pareto boundary}, or \textit{Pareto front}.

For calculating the Pareto boundaries, we have used the deep neuroevolution strategy described in~\cite{UberDeepNeuroevolution2018}. Using genetic algorithms~\cite{Deb11}, we have evolved the weights of a population $\Phi$, where $\Theta$ is a solution vector containing the weights of an individual $\varphi (\Theta)$. 

The first training step consists of running a forward pass through $\Phi$, obtaining values for the multi-objective fitness vector $\vec{L}$. After completing the forward passes, we use the tournament selection algorithm~\cite{Deb11} to select the elite individuals, that is, the individuals located on the Pareto boundary in objective space $\vec{L}$. With each training iteration, a new Pareto front is calculated. The tournament selection algorithm assures that a number of elite individuals carry on to the next training generation unmodified. For exploring the decision space $\vec{S}$, we use mutation and a uniform crossover between $2$ individuals. Both mutation and crossover operations are randomly applied with a $50\%$ probability. 



During training, each generation of individuals is subjected to additive Gaussian noise $\sigma$:

\begin{equation}
	\Theta := \Theta + \sigma,
	\label{eq:added_gaussian_noise}
\end{equation} 

\noindent where $\sigma \in [-3,3]$ was chosen with respect to the sigmoid function used for activating the neurons in the deep networks. The additive Gaussian noise aims to not oversaturate the weights by reaching maximum or minimum values in the sigmoid activation function.

\begin{figure}
	\centering
	\begin{center}
		\includegraphics[scale=0.9]{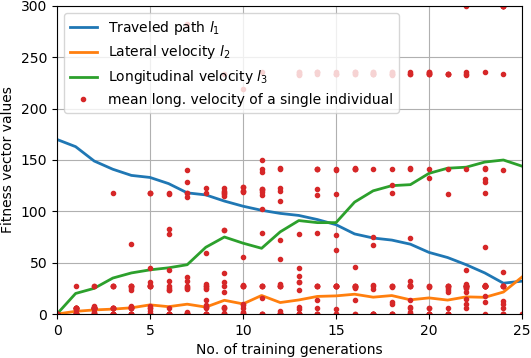}
		\vspace{-0.8em}
		\caption{\textbf{Evolution of the fitness vector during training}. With each training generation, the traveled path decreases, while the longitudinal velocity increases. The lateral velocity increases together with the longitudinal velocity, but with a much smaller gradient, meaning that the vehicle is learning to avoid hazardous motions and passenger discomfort, although the longitudinal velocity is high. The red dots show the mean fitness value for the longitudinal velocity at each training generation.}
        \label{fig:fitness_vector_evolution_during_training}
	\end{center}
	\vspace{-0.7em}
\end{figure}

The new population is evaluated and the process repeats itself for a given number of training generations, at the end of which the final Pareto front $\Phi^*$ is obtained.
The evolution of the fitness vector in Eq.~\ref{eq:multiobjective_loss} during training can be seen in Fig.~\ref{fig:fitness_vector_evolution_during_training}. Details on Pareto optimization using evolutionary methods can be found in~\cite{Deb11}.

\section{Experiments}
\label{sec:experiments}

\subsection{Experimental Setup Overview}
\label{sec:experiments_overview}

\begin{figure}
	\centering
	\begin{center}
		\includegraphics[scale=0.42]{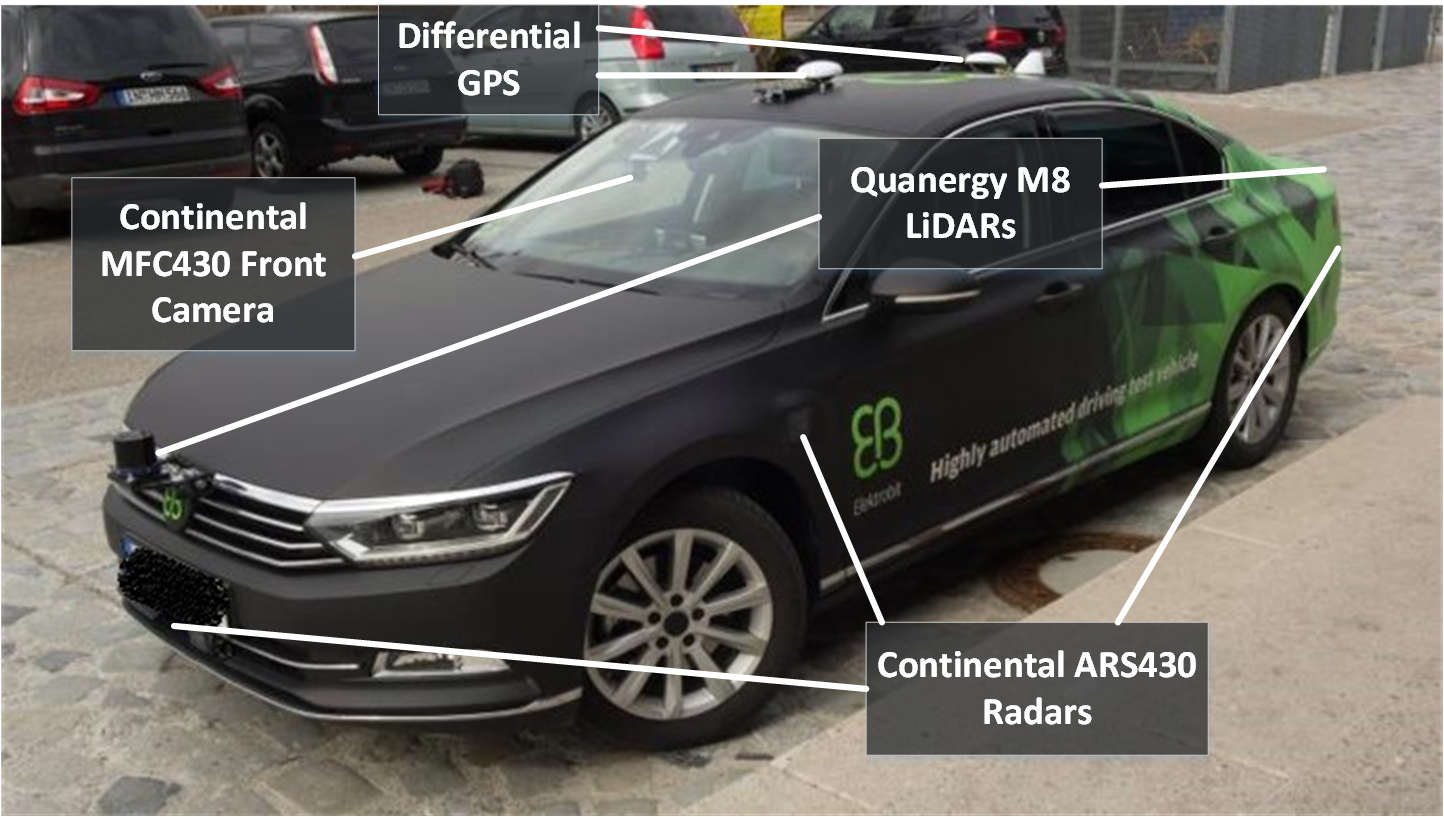}
		\vspace{-0.3em}
		\caption{\textbf{Autonomous test vehicle used for real-world data acquisition.} The car is equipped with a front Continental MFC430 camera, two front and rear Quanergy M8 Lidars and six front, rear and side Continental ARS430 radars.}
        \label{fig:eb_car}
	\end{center}
	\vspace{-2.5em}
\end{figure}

The performance of NeuroTrajectory was benchmarked against the baseline Dynamic Window Approach (DWA)~\cite{Fox_Dynamic_Window_Approach_1997}, as well as against End2End supervised learning~\cite{NVIDIA_End2End_Learning}.

We have tested the NeuroTrajectory algorithm in two different environments: \textit{I}) in the GridSim simulator and \textit{II}) with the full scale autonomous driving test car from Fig.~\ref{fig:eb_car}. This resulted in $20km$ of driving in GridSim and over $100km$ of trajectory estimation with our neuroevolutionary system on real-world roads. The three variants had to solve the same optimization problem illustrated in Fig.~\ref{fig:problem_description}, which is to calculate a trajectory for safely navigating the driving environment, without performing the motion control task.

The Dynamic Window Approach (DWA)~\cite{Fox_Dynamic_Window_Approach_1997} is an online collision avoidance strategy for mobile robots, which uses robot dynamics and constraints imposed on the robot's velocities and accelerations to calculate a collision free path in the 2D plane. We have implemented DWA based on the Robot Operating System (ROS) DWA local planner.

In the case of End2End learning, we have mapped sequences of OGs directly to the discrete driving commands of the vehicle, as commonly encountered in deep learning based autonomous driving~\cite{Jaritz_ICRA2018}. The network topology is based on the same deep network architecture from Fig.~\ref{fig:neural_network_diagram}, having the same configuration for the number of layers and processing units. The difference is that the final layer of an LSTM branch is configured to output discrete steering commands (turn left, turn right, accelerate, decelerate), instead of a continuous state trajectory.

\subsection{Real-World and Synthetic Training Datasets}
\label{sec:real_dataset}

Real-world training data has been collected by driving on several types of roads, using the Volkswagen Passat test car shown in Fig.~\ref{fig:eb_car}. The vehicle is equipped with a front Continental MFC430 camera, two front and rear Quanergy M8 Lidars and six front, rear and side Continental ARS430 radars.

The sensory data streams are fused into an occupancy grid of size $125m \times 125m$, having a $0.25m$ cell resolution. Each data sample is acquired at time intervals ranging from $50ms$ to $90ms$ for one cycle. In total, approximately $63.000$ samples were acquired in different driving scenarios, such as country roads, highways, city driving, T junctions, traffic jams and steep curves.

Synthetic data is acquired using GridSim~\cite{Trasnea_IRC2019}~\footnotetext{GridSim has been released as open-source software, available on GitHub. Additional information is available at \url{www.rovislab.com/gridsim.html}}. GridSim is an autonomous driving simulation engine that uses kinematic models to generate synthetic occupancy grids from simulated sensors. It allows for multiple driving scenarios to be easily represented and loaded into the simulator. The virtual traffic participants inside GridSim are generated by sampling their behavior from a database of prerecorded driving trajectories. The simulation parameters for seamless and inner-city scene types are given in Table~\ref{tab:simulatoin_parameters}. The testing datasets are different from the training ones, for both the simulation and real-world experiments.

\begin{table}
	\centering
	\begin{tabular}{lll}
		\hline
		\textbf{Description} & \textbf{Seamless} & \textbf{Inner-city} \\
		\hline
		Number of traffic participants & 10 & 20 \\
		Maximum speed/acceleration & $13.88 m/s$ & $8.33 m/s$ \\
		Maximum acceleration & $2 m/s^2$ & $2 m/s^2$ \\
		Minimum speed & $4.16 m/s$ & $2.77 m/s$ \\
		Percentage of straight road & $60\%$ & $45\%$ \\
		Mean curve radius & $55^{\circ}$ & $81^{\circ}$\\
		\hline
	\end{tabular}
	\caption{GridSim simulation parameters.}
	\label{tab:simulatoin_parameters}
	\vspace{-2.3em}
\end{table}

\subsection{Experiment I: Simulation Environment}
\label{sec:experiments_I}

The first set of experiments compared three algorithms (DWA, End2End and NeuroTrajectory) over $20km$ of driving in GridSim~\cite{Trasnea_IRC2019}.

\begin{figure}
	\centering
	\begin{center}
		\includegraphics[scale=0.95]{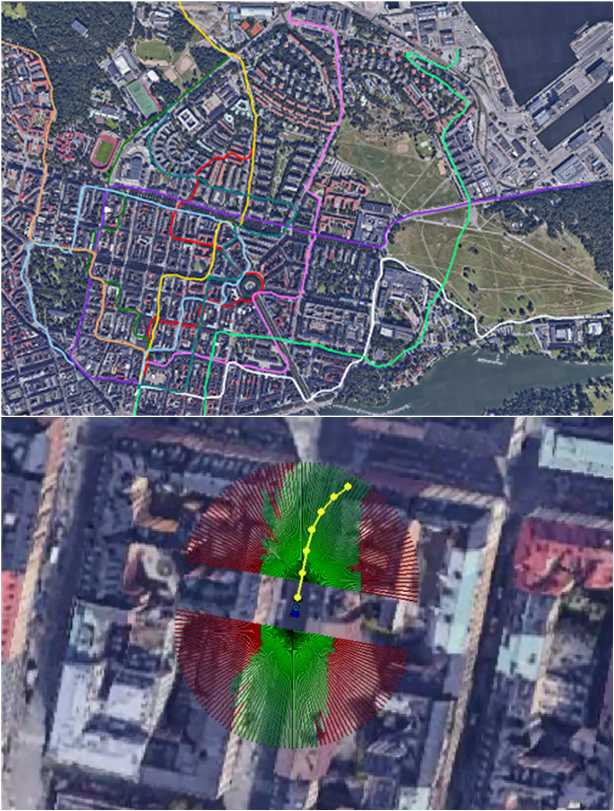}
		\vspace{-0.3em}
		\caption{\textbf{GridSim evaluation routes}. The virtual test field is defined on $30 km^2$ of the Stockholm inner-city area (upper image). The ego-vehicle perceives the driving environment using an OG structure; the predicted trajectory is shown in yellow (lower image).}
        \label{fig:stockholm_routes}
	\end{center}
	\vspace{-2.3em}
\end{figure}

The task is to reach a given goal location from a starting position, while avoiding collisions and driving at desired speed. The evaluation routes, illustrated in Fig.~\ref{fig:stockholm_routes}, where defined within a $30 km^2$ inner-city area, mapped in GridSim from a downtown area of Stockholm.

As performance metric, we use the Root Mean Square Error (RMSE) between the estimated and a human driven trajectory in the 2D driving plane:

\begin{equation}
	\vspace{-0.2em}
	RMSE = \sqrt{ \frac{1}{\tau_o} \sum^{\tau_o}_{t=1} \left[ (\hat{p}^{<t>}_x - p^{<t>}_x)^2 + (\hat{p}^{<t>}_y - p^{<t>}_y)^2 \right] },
	\label{eq:rmse}
\end{equation}

\noindent where $\hat{p}^{<t>}_x$, $p^{<t>}_x$, $\hat{p}^{<t>}_y$ and $p^{<t>}_y$ are set-points on the estimated and human driven trajectories, respectively, along the $x-y$ driving plane. $\tau_o = 5$ is the prediction horizon.

We have implemented an End2End supervised learning system which predicts the vehicle's steering angle, discretized with a $3^{\circ}$ resolution. Given the predicted steering angle, we define a line segment centered on the ego-vehicle. Ideally, the angle between the line segment and the line defined by the first two set-points of the reference trajectory should be zero. As performance metric, we measure the RMSE between the reference points and their closest set-points on the line segment, for $\tau_o = 5$.


The performance evaluation of the benchmarked algorithms is summarized in Table~\ref{tab:accuracy_results}, for the different types of roads and traffic environments encountered in the synthetic testing database. We show the mean ($\bar{e}_x$, $\bar{e}_y$) and maximum ($\max(e_x)$, $\max(e_y)$) position errors, together with the RMSE metric from Eq.~\ref{eq:rmse}. The errors are measured for trajectory estimation in GridSim, as well as for predictions on real-world highway and inner-city driving data.

\begin{table*}
	\centering
	\begin{tabular}{|l|ccc|ccc|ccc|}
		\hline
		\textbf{Metric} & \textbf{DWA} & \textbf{End2End} & \textbf{Ours} & \textbf{DWA} & \textbf{End2End} & \textbf{Ours} & \textbf{DWA} & \textbf{End2End} & \textbf{Ours} \\
		$[m]$ & GridSim & GridSim & GridSim & Highway & Highway & Highway & Inner-city & Inner-city & Inner-city \\
		\hline
		$\bar{e}_x$ & 1.58 & 3.51 & \textbf{1.03} & 0.86 & 3.55 & \textbf{0.31} & 4.49 & 12.85 & \textbf{3.57} \\
		$\max(e_x)$ & 6.09 & 10.81 & \textbf{4.91} & 2.86 & 4.69 & \textbf{1.55} & 6.84 & 15.18 & \textbf{8.51} \\
		$\bar{e}_y$ & 2.16 & 4.83 & \textbf{1.82} & 1.51 & 2.09 & \textbf{0.84} & 1.72 & 9.03 & \textbf{1.08} \\
		$\max(e_y)$ & 7.33 & 11.72 & \textbf{3.86} & \textbf{2.04} & 8.14 & 2.92 & 12.84 & 17.33 & \textbf{10.88} \\
		$RMSE$ & 2.68 & 5.97 & \textbf{2.09} & 1.74 & 4.12 & \textbf{0.90} & 4.81 & 15.71 & \textbf{3.73} \\
		\hline
	\end{tabular}
	\caption{Comparative accuracy results obtained in GridSim, as well as on real-world highway and inner-city driving data.}
	\label{tab:accuracy_results}
	\vspace{-2.3em}
\end{table*}

\subsection{Experiment II: Real-World Data}
\label{sec:experiments_II}

\begin{figure}
	\centering
	\begin{center}
		\includegraphics[scale=0.62]{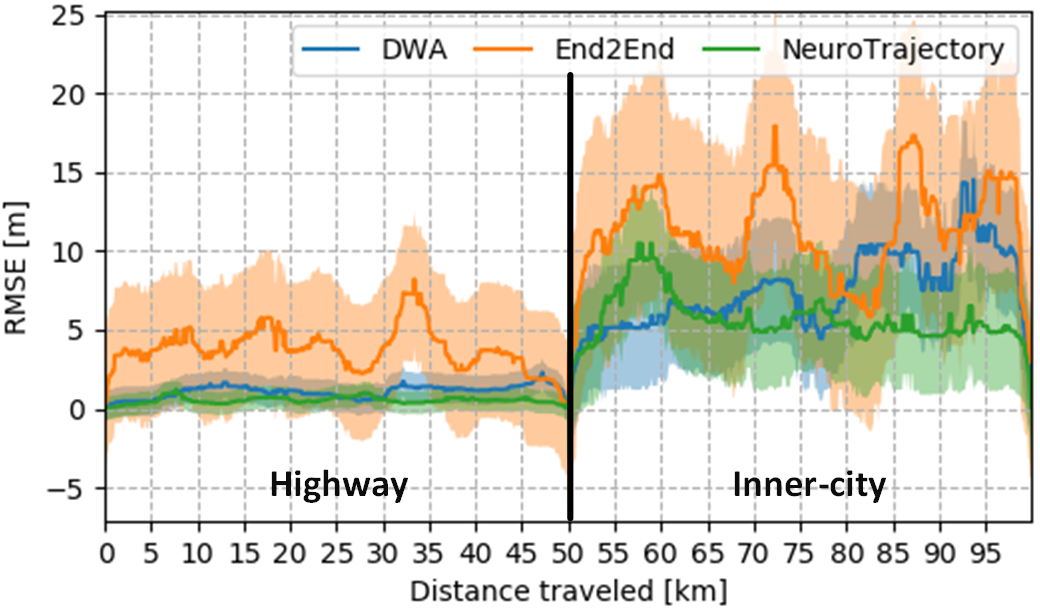}
		\vspace{-0.5em}
		\caption{\textbf{RMSE between estimated and human driven trajectories in real-world highway and inner-city testing scenarios}. The solid lines indicate the position error, calculated as the RMSE from Eq.~\ref{eq:rmse}. The shaded region indicates the standard deviation. The position errors are higher for inner-city scenes, then in the case of highway driving. NeuroTrajectory achieves the lowest error for both testing scenarios.}
        \label{fig:position_error}
	\end{center}
	\vspace{-2.5em}
\end{figure}

\begin{figure}
	\centering
	\begin{center}
		\includegraphics[scale=1.0]{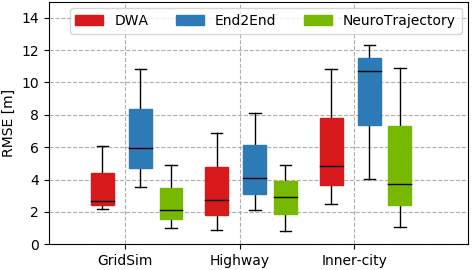}
		\vspace{-1.0em}
		\caption{\textbf{Median and variance of RMSE for the three testing scenarios}. The errors in simulation and highway driving are similar. The unstructured nature of inner-city driving introduces higher errors, as well as a higher RMSE variance.}
        \label{fig:accuracy_results}
	\end{center}
	\vspace{-2em}
\end{figure}

Using the test car in Fig.~\ref{fig:eb_car}, we have performed experiments on sequences of occupancy grids acquired on $50km$ of highway and $50km$ of inner-city driving. As in the simulation experiment, the path driven by the human driver was considered as ground truth for evaluation. The position errors are shown in Fig.~\ref{fig:position_error} for both scene types. Fig.~\ref{fig:accuracy_results} shows the median and variance of the RMSE recorded in the three testing scenarios. The standard deviation and variance have been calculated based on the mean RMSE for multiple estimated trajectories provided by a single trained instance of the deep network from Fig.~\ref{fig:neural_network_diagram}. Fig.~\ref{fig:accuracy_results} indicates that the prediction performance remains similar in the case of simulation and highway driving, while the prediction error increases for inner-city scenes, for all three benchmarked algorithms. DWA has the closest errors to NeuroTrajectory, making it still a good candidate for trajectory estimation using non-learning methods.

Errors recorded in the inner-city scenario are higher than in the case of highway driving. This is mainly due to the highly unstructured scenes encountered when driving in a city, as opposed to highway driving. NeuroTrajectory achieved the lowest error measures both in simulation, as well as in real-world testing, except for the maximum longitudinal error $\max(e_y)$ during real-world highway driving, which was smaller for DWA.

In our experiments, DWA behaved better than End2End mostly due to the structure of the OG input data, which is more suited for grid-based search algorithms. This makes DWA strictly dependent on the quality of the OGs, without having the possibility to apply such classical methods directly on raw data. Paradigms such as NeuroTrajectory, or End2End, would scale without any customizations on raw sensory information (e.g. video streams, radar, Lidar, etc.). Additionally, the jittering effect of End2End can be a side effect produced by the discrete nature of its output. However, NeuroTrajectory is able to combine both worlds and obtain a stable state prediction of the ego-vehicle along a given time horizon. We believe that in the long run, learning based approaches will produce better results than traditional methods, like DWA. This improvement would come from training on additional data, including a larger amount of corner cases.

\section{Conclusions}
\label{sec:conclusions}

In summary, this paper introduces NeuroTrajectory, which is a neuroevolutionary approach to local state trajectory learning from sequences of occupancy grids. The learning procedure is based on multi-objective Pareto optimization, encoding learned trajectories into the weights of a perception-planning deep neural network. For training the network, we use real-world and synthetic occupancy grids, together with their corresponding driving commands, recorded from a human driver. The synthetic data is generated using our GridSim simulator. Based on performance evaluation against the DWA baseline and End2End supervised learning, we conclude that the proposed neuroevolutionary approach is a valid candidate for local trajectory estimation in autonomous driving. As future work, we would like to extend the validation of our approach on additional driven kilometers, as well as on different driving conditions.

\bibliographystyle{IEEEtran}
\bibliography{references}








\end{document}